\pdfoutput=1

\documentclass[11pt]{article}

\usepackage[final]{acl}

\usepackage{times}
\usepackage{latexsym}

\usepackage{microtype}
\usepackage{hyperref}
\usepackage{url}
\usepackage{enumitem}
\usepackage{multicol}
\usepackage{multirow}
\usepackage{CJKutf8}
\usepackage{amsmath}
\usepackage{amssymb}
\usepackage{siunitx}
\usepackage{floatflt}
\usepackage{graphicx}
\usepackage{wrapfig}
\usepackage{lipsum}
\usepackage{dsfont}
\usepackage{bbm}
\usepackage{algorithm}
\usepackage{algorithmicx}
\usepackage{algpseudocode}
\usepackage{microtype}
\usepackage{graphicx}
\usepackage{subfigure}
\usepackage{booktabs} 
\usepackage{pifont}  
\usepackage{graphicx}  
\usepackage{subcaption} 
\usepackage{hyperref}
\usepackage{subcaption}

\usepackage[most]{tcolorbox}
\newtcolorbox{alprompt}[1]{
        boxrule = 1pt,
        fontupper = \small\tt,
        fonttitle = \bf\color{black},
        arc = 2pt,
        rounded corners,
        colframe = black,
        colbacktitle = white!97!yellow,
        colback = white!97!yellow,
        title = #1,
}
\definecolor{darkgreen}{rgb}{0.0, 0.5, 0.0}
\definecolor{darkgray}{gray}{0.4}
\definecolor{maroon}{rgb}{0.5, 0.0, 0.0}
\definecolor{navy}{rgb}{0.0, 0.0, 0.5}
\definecolor{teal}{rgb}{0.0, 0.5, 0.5}

\definecolor{deepblue}{RGB}{41, 128, 185}

\definecolor{mylightgreen}{RGB}{144,238,144}
\definecolor{mylightblue}{RGB}{173,216,230}

\definecolor{outerboxcolor}{gray}{0.90} 
\definecolor{innerboxcolor}{rgb}{1,1,1}

\definecolor{nred}{RGB}{196, 38, 11}
\definecolor{ngreen}{RGB}{18, 141, 21}
\definecolor{nblue}{RGB}{41, 52, 190}
\newcommand{\finalcopy}[1]{\textcolor{blue}{#1}}

\newcommand{\tabincell}[2]{\begin{tabular}{@{}#1@{}}#2\end{tabular}} 

\algnewcommand{\LeftComment}[1]{\Statex \(\triangleright\) #1}

\usepackage{array}
\usepackage{amsmath}
\usepackage{mathtools}
\usepackage{amsthm}
\usepackage{arydshln}
\usepackage[capitalize,noabbrev]{cleveref}
\usepackage{adjustbox} 
\usepackage{enumitem}
\usepackage{longtable}

\usepackage[T1]{fontenc}

\usepackage[utf8]{inputenc}

\usepackage{microtype}

\usepackage{inconsolata}

\usepackage{graphicx}

\title{WebEvolver: Enhancing Web Agent Self-Improvement\\ with Co-evolving World Model}

\author{
Tianqing Fang, Hongming Zhang, Zhisong Zhang, Kaixin Ma,  Wenhao Yu, \\
\textbf{ Haitao Mi, Dong Yu}\\
Tencent AI Lab\\
\texttt{tianqfang@tencent.com}
}

\begin{document}
\maketitle
\begin{abstract}

Agent self-improvement, where agents autonomously train their underlying Large Language Model (LLM) on self-sampled trajectories, shows promising results but often stagnates in web environments due to limited exploration and under-utilization of pretrained web knowledge.
To improve the performance of self-improvement, we propose a novel framework that introduces a co-evolving World Model LLM.
This world model predicts the next observation based on the current observation and action within the web environment.
The World Model serves dual roles: 
(1) as a virtual web server generating self-instructed training data to continuously refine the agent's policy, 
and (2) as an imagination engine during inference, enabling look-ahead simulation to guide action selection for the agent LLM. 
Experiments in real-world web environments (Mind2Web-Live, WebVoyager, and GAIA-web) show a 10\% performance gain over existing self-evolving agents, demonstrating the efficacy and generalizability of our approach, without using any distillation from more powerful close-sourced models\footnote{Code is available at \url{https://github.com/Tencent/SelfEvolvingAgent}}. 
\end{abstract}

\section{Introduction}

\begin{figure}[t]
    \centering
    \includegraphics[width=0.48\textwidth]{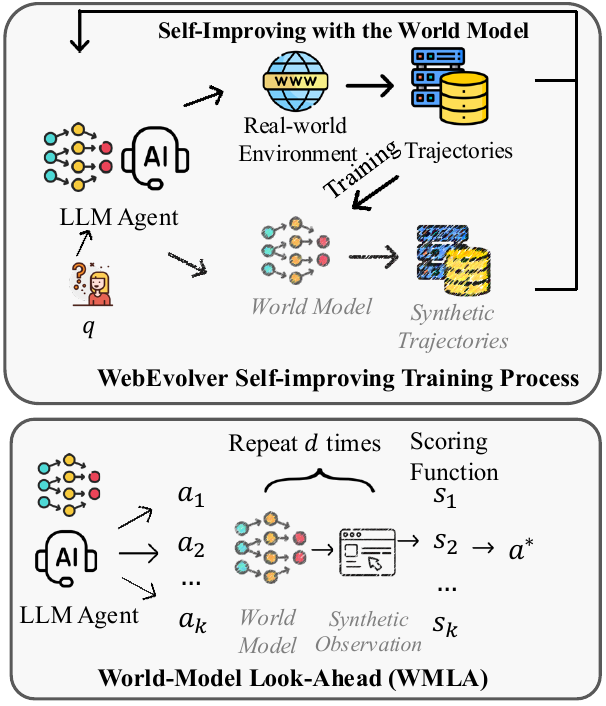} 
    \caption{Overview of WebEvolver – A Self-Improving Framework with World-Model Look-Ahead.
    Our framework co-trains a world model with the agent to predict next-step observations based on current states and actions. The world model serves as a virtual web engine, which generates synthetic trajectories for policy training and enables look-ahead planning to select optimal actions during inference.}
    \label{fig:overview}
\end{figure}

Autonomous agents, especially Web agents operating in online environments, play a crucial role in automating complex tasks, advancing progress towards artificial general intelligence~\citep{deep_research, manus, ui_tars, openmanus2025}.
The capabilities of these agents stem from two key components, the design of the system, which facilitates accessing and processing abundant information from the web, and the agent foundation language model itself, which is typically a (Multimodal) Large Language Model (LLM) that generates actions based on the provide context.

Recent work in \textbf{agent self-improvement} refines LLM-based agents through iterative cycles of autonomous interaction: agents generate actions, collect behavioral trajectories, and are fine-tuned on this self-collected data after rejection sampling~\citep{godal_agent, BAGEL, LLM_agent_can_self_improve, rest_meets_react, openwebvoyager, AgenGym}. While this bootstrapping reduces reliance on human-labeled data, performance eventually plateaus~\citep{B_STAR}.

This stagnation arises from two main bottlenecks. First, exploration diversity declines as the agent overfits to familiar trajectories, limiting discovery of novel states~\citep{openwebvoyager}. Second, although inference-time exploration methods~\citep{AgentTreeSearch, webpilot, LATS, AgentQ, Exact} have the potential to provide diverse trajectories, they require costly real-world interactions for marginal gains. 
On the other hand, simulation or imagination-based approaches~\citep{webdreamer, agent_world_model} typically offer only one/two-step look-ahead, lacking coherent multi-step rollouts.

To address these limitations, we propose integrating a \textbf{Co-evolving World Model} into the self-improvement loop to enable better multi-step trajectory synthesis and look-ahead. 
Our world model is a language model trained to predict the next observation (web page) given the current state and an attempted action.
Our key insight is that LLMs, pretrained on vast web content~(e.g., Llama-3; \citealp{llama3}), inherently encode a structured understanding of website dynamics, user intents, and task workflows. 
We fine-tune it on trajectories collected during agent-environment interactions, allowing it to evolve alongside the agent to provide better simulation results.


As a \textit{virtual web server}, The World Model serves two roles : (1) it generates diverse, self-instructed training trajectories by simulating interactions with unseen web environments, mitigating exploration bottlenecks by exposing the agent to a wider range of scenarios than real interactions alone. While the World Model may produce some hallucinated (i.e., non-realistic) web states, this is not critical during training, as the agent's goal is to learn flexible action prediction, even under noisy circumstances. 
(2) during inference, the World Model performs multi-step look-ahead simulations~\citep{SLA}, allowing the agent to evaluate possible actions without costly real-world trials. 
This dual mechanism grounds self-improvement in both real and model-based interactions, ensuring sustained adaptability while reducing reliance on expensive environment interactions.

We validate our framework on real-world, open-domain web environments, including Mind2Web-Live~\citep{webcanvas}, WebVoyager~\citep{webvoyager}, GAIA-web~\citep{GAIA}, and SimpleQA~\citep{simpleqa}\footnote{We adapt this dataset to search queries on the internet}. Experiments show a 10\% performance improvement over the self-evolving baseline OpenWebVoyager~\citep{openwebvoyager}, with notable gains on complex and unseen tasks.

Our main contributions are:
\begin{enumerate}
\item Introducing the co-evolving world model for self-improving web agents, enabling diverse training data generation and low-cost multi-step action search.
\item Providing empirical evidence that world-model-guided self-improvement enhances agent performance and adaptability in open-domain settings, with minimal human supervision and no distillation from stronger LLMs.
\end{enumerate}

This work highlights the importance of integrating dynamic world models into agent frameworks to overcome the limitations of purely data-driven self-training.


\begin{figure*}[t]
    \centering
    \includegraphics[width=\textwidth]{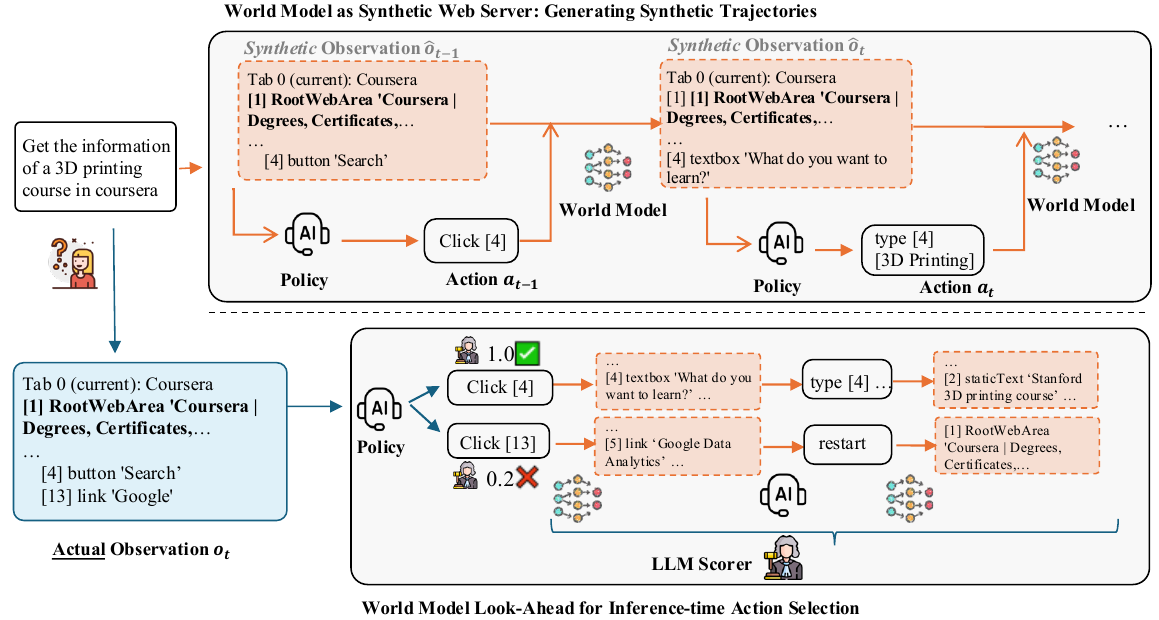} 
    \caption{An illustration of the World Model trajectory synthesizing process and World Model Look-ahead for inference-time action selection.}
    \label{fig:method}
\end{figure*}

\section{Related Work}

\paragraph{Web Agent}
Recent advances in web agents leverage (multimodal) large language models as their backbone \citep{llama3, leopard, gpt4, claude37}, enabling reasoning through frameworks like ReAct~\cite{yaoreact}, MCP~\citep{MCP}, and cognitive kernel~\citep{cognitive_kernel}. 
These agents are evaluated on benchmarks such as WebShop~\citep{webshop}, Mind2Web~\citep{mind2web}, WebArena~\citep{webarena}, VisualWebArena~\citep{visualwebarena}, WebVoyager~\citep{webvoyager}, WebWalker~\citep{WebWalker}, and MMInA~\citep{MMInA}.
Besides applying off-the-shelf LLMs, there are data scaling efforts like Explorer \citep{Explorer}, NNetNav \citep{NNetNav}, and InSTA \citep{DBLP:journals/corr/abs-2502-06776} enhance the training of LLMs. 
Inference-time optimization techniques, including AgentTreeSearch \citep{AgentTreeSearch}, Monte-Carlo Tree Search \citep{AgentQ, Exact, LATS, webpilot}, and Reflexion \citep{Reflexion}, further improve decision-making.
More recently, the development of web agents focus on multi-step Deep Research agent and the training of agent foundation models, such as WebThinker~\cite{li2025webthinker}, WebDancer~\cite{wu2025webdancer}, WebSailor~\cite{li2025websailor}, WebShaper~\cite{tao2025webshaper}, Cognitive Kernel-Pro~\cite{cognitivekernel-pro}, MiroFlow~\cite{MiroFlow2025}, and so on.

\paragraph{Agent Self-Improvement} Beyond using off-the-shelf LLMs as policy models or fine-tuning via imitation learning from powerful LLM trajectories, recent work explores bootstrapping agent LLMs with open-source models~\citep{rest_meets_react, LLM_agent_can_self_improve}, building on advances in self-improving LLM reasoning~\citep{self_instruct, STAR, B_STAR}.
BAGEL~\citep{BAGEL}, OpenWebVoyager~\citep{openwebvoyager}, and Self-Improved Agents~\citep{LLM_agent_can_self_improve} explored iterative exploration-feedback-optimization cycles, where agents refine their policies by learning from high-quality trajectories in real-world or simulated web environments. 
To enhance self-improvement, G"{o}del Agent~\citep{godal_agent} enables agents to dynamically modify their logic and accumulate skills across diverse tasks.
WebCoT~\cite{hu2025webcot} study cognitive behavior injection to the Chain-of-thought of Agent Foundation Model to improve the reasoning ability of the agents.
\cite{zhang2025enhancingwebagentsexplicit} explores bootstrapping the ability of backtracking in web agent tasks.
AgentQ~\citep{AgentQ} and ReST+ReAct~\citep{rest_meets_react} combine reinforcement learning and preference optimization, enabling agents to learn from both successes and failures and improving robustness in multi-step reasoning.
While reinforcement learning is promising for self-improvement, real-world, evolving websites pose challenges: environmental uncertainty can lead to inconsistent evaluations of the same action, making it difficult for agents to reliably assess and improve their performance.
In addition, while analogous to reinforcement learning (RL), our iterative Supervised Fine-Tuning (SFT) paradigm (with binary trajectory selection) differs from end-to-end RL as it optimizes LLM loss rather than reward functions directly. We collect the agent SFT data first and then use standard LLM SFT framework to optimize the LLM, which is why similar paradigms were termed as agent self-improvement rather than RL.

\paragraph{World Models} 

World models have evolved from their reinforcement learning origins \citep{world_model} to become powerful tools for agent reasoning~\cite{DBLP:journals/corr/abs-2408-14837, DBLP:conf/nips/AlonsoJMKSPF24, DBLP:journals/corr/abs-2305-14223}. Recent approaches leverage large language models (LLMs) as implicit world models, enabling agents to simulate and plan through complex tasks. For general reasoning, RAP~\citep{RAP_world_model} demonstrates how LLMs can serve dual roles as both world models and reasoning agents, using Monte Carlo Tree Search to explore future states. Similarly, WKM~\citep{agent_world_model} shows that structured world knowledge can be distilled from trajectories to guide agent planning.
In web environments, methods like WebDreamer~\citep{webdreamer} and WMA~\citep{web_agent_world_model} adapt this paradigm by using LLMs to predict action outcomes through natural language simulations.
However, these approaches remain limited by their reliance on off-the-shelf LLMs.
In detail, both WebDreamer and WMA works at inference time. Their approaches essentially create sophisticated chain-of-thought pipelines where the world model serves as intermediate information for static GPT-4o. Despite that WMA uses a finetuned world model instead of GPT-4o, eventually the inference-time algorithm is still a prompting pipeline. Neither method actually uses the world model to improve the agent itself through training.
Besides, despite WMA also has multi-step simulation (depth up to 3), it is only used for inference-time action selection. Instead, our approach generate trajectories using world model with depth up to 7, and the full trajectories are used for training instead of only for prompting to select best actions.

Our work advances beyond these limitations by co-learning a dedicated world model during agent self-improvement. This enables genuine multi-step trajectory synthesis and look-ahead planning, providing a more robust foundation for interactive decision-making than current prompt-based approaches.

\section{Method}

In this section, we introduce the WebEvolver, a co-learning framework of World Model and Agent Policy model (Figure \ref{fig:method}). 

\subsection{Problem Formulation}

The web agent task is formulated as a Partially Observable Markov Decision Process (POMDP) 
$(\mathcal{S}, \mathcal{A}, \mathcal{O}, \mathcal{T}, \mathcal{R})$,
where the agent receives a natural language query $q$ requiring multi-step web interaction under the environment. 
The state space $\mathcal{S}$ represents the complete web environment, 
while the observation space $\mathcal{O}$ is limited to visible elements. At each time step $t$: $o_t = \Omega(s_t)$, where $\Omega$ is a function extracting visible contents like $(\text{URL}, \text{Web Elements})$ from the current state $s_t$. 
$\mathcal{A}$ represents the whole action space, which, in our case we include \texttt{click}, \texttt{type}, \texttt{goback}, \texttt{scroll down/up}, and \texttt{stop}, as the atomic web operations.
$\mathcal{T}$ represents the deterministic transition function that executes browser operations to advance the state.
The agent's policy $\pi(o_t, q) \rightarrow a_t$ generates actions that produce trajectories $\tau = \{(o_1,a_1),\ldots,(o_t,a_t)\}$, 
with final rewards computed through self-assessment $\hat{r}(\tau, q) \in [0,1]$. 

Given a task query $q$ and target website $w$, we initialize the web environment and get the first observation $o_1 \in \mathcal{O}$. 
We follow the settings in Cognitive Kernel~\citep{cognitive_kernel} and use accessibility tree to represent the elements in $o_t$.
Using an LLM as agent policy model parameterized by $\theta$, we generate chain-of-thoughts $h_t$ and actions $a_t$ at time step $t$:


\begin{equation}
(h_t, a_t) \sim \pi_\theta(\cdot|I, q, o_{1:t}, h_{1:t-1}, a_{1:t-1})
\end{equation}

where $I$ contains system instructions. 
The transition function $\mathcal{T}$ executes actions on the environment:

\begin{equation}
s_{t+1} = \mathcal{T}(s_t, a_t), \ o_{t+1} = \Omega(s_{t+1})
\end{equation}

The complete trajectory is $\tau = (o_1, h_1, a_1, \dots, o_T, h_T, a_T)$, where $T$ denotes the total number of navigation steps.

\subsection{Agent Self-Improvement}

In this subsection, we introduce the self-improvement of a backbone agent foundation model, denoted as $\mathcal{M}$, and the corresponding policy function is denoted as $\pi_{\mathcal{M}}$.

\paragraph{Trajectories Collection}
We employ $\mathcal{M}$ to sample actions based on an input query $q$, 
which are then used to collect web navigation trajectories. 
We use $\mathcal{M}$ as the agent foundation model to power Cognitive Kernel, which interacts with web environments. 
The agent observes the last $k$ steps, represented as webpage accessibility trees, to inform its actions.

For each query $q \in \mathcal{Q}$, a trajectory $\tau_i$ is sampled from the policy $\pi_{\theta_M}(\tau \mid I, q)$. 
To prevent performance degradation from too long contexts, we clip the trajectory history $c_t$ when $t - 1 > k$ by keeping only the latest observations. The thoughts and actions are kept as they contain some compressed information about the history.

\begin{align} \label{equation:c-clip}
    c_t^{\text{clip}} =& (h_1, a_1, h_2, a_2, \dots, h_{t-k}, a_{t-k},  \nonumber \\
    & o_{t-k+1}, h_{t-k+1}, a_{t-k+1}, \dots, o_{t-1}),
\end{align}

such that the new actions are generated with the following function:

\begin{equation} \label{equation:gpt-4o-reasoning}
    (h_t, a_t) \sim \pi_{\theta_M}(\cdot \mid I, q, c_t^{\text{clip}}).
\end{equation}

Notably, we retain the \textbf{thought} and \textbf{action} at each step to preserve the full reasoning chain while avoiding context overload.
Then, rejection sampling is conducted to keep those trajectories that are successfully finished, using an automatic evaluation method $\hat{r}(\tau, q)$. 

\paragraph{Iterative Optimization}

At the $i$-th iteration of the self-improvement, we denote the collected trajectories after rejection sampling as $D_{i}$.
We aim to maximize the following objective function:

\begin{align}\label{equation:J-bc}
    \mathcal{J}(\theta) = \mathbb{E}_{(q,\tau) \sim D_{\text{i}}} \sum_{t=1}^T \Big[\log \pi_\theta(a_t | q, c_t^{\text{clip}'}, h_t) \nonumber  \\
    + \log \pi_\theta(h_t | q, c_t^{\text{clip}'})\Big],
\end{align}

After acquiring the new policy model $\mathcal{M}_i$, it is used to sample trajectories from the query set $\mathcal{Q}$ again. The newly successful trajectories are then appended to $D_{i}$ to form a new training dataset $D_{i+1}$ to perform the next round of optimization.

\subsection{WebEvolver}

In this subsection we introduce the co-learning/co-training world model, and how to use it for trajectory synthesizing and inference-time look-ahead. An illustration figure is presented in Figure \ref{fig:method}.

\paragraph{Co-learning World Model}
The world model is a language model that simulates the next observation $\hat{o}_{t+1}$ conditioned on both the current webpage's accessibility tree ($o_t$) and a formatted action string ($a_{t-1}$), thereby predicting state transitions.
We learn a world model LLM $\mathcal{M}_w$ using the collected trajectory during self-improvement. 

From the a collected trajectory $\tau = \{(o_0,a_0),\ldots,(o_t,a_t)\}$, we can convert it to a world modeling trajectory $\tau_w = \{o_0, (a_0, o_1),\ldots,(a_{t-1},o_t)\}$, such that the objective of world model is to predict the next observation $o_t$ conditioned on the scheduled action $a_{t-1}$ and previous observations. 
Similar with the trajectories in agent policy model, we truncate the history observations to avoid performance degrade on long contexts. 
Here, we simply use the latest observation as history.
Besides, we distill some rationales using the original base LLM $\mathcal{M}$ about the logic of the transition function $\mathcal{T}$ to help the generation of the next webpage. 
Such chain-of-thoughts at step $t$ is denoted as $h^w_t$.
We do not omit the action and thoughts to make the world model aware of some of the previous information and the depth of the trajectory.
\vspace{-1em}

\begin{equation}
c^w_t = (a_1, h^w_1, \dots, a_{t-2}, h^w_{t-2}, o_{t-1}, a_{t-1}), 
\end{equation}

Such that the next webpage observation $o_t$ is generated with the following function, where $\theta_w$ is the parameters of $\mathcal{M}_w$.

\begin{equation}
o_t \sim \pi_{\theta_w}(\cdot|I_w, c^w_t)
\end{equation}

The world model is then optimized using the latest iteration of collected trajectories.
\vspace{-1em}
\begin{align}\label{equation:J-wm}
    \mathcal{J}(\theta_w) = \mathbb{E}_{\tau_w \sim D_{\text{i}}} \sum_{t=1}^T \Big[\log \pi_{\theta_w}(\finalcopy{o}_t |  c^w_t, h^w_t)  \nonumber \\
    + \log \pi_{\theta_w}(h^w_t | c^w_t)\Big],
\end{align}

\paragraph{Trajectory Synthesis}

We can use an agent policy model $M_{i}$ and a world model $M_{w}$ to perform synthetic trajectory generation, enabling us to scale up the training data without interacting with the real web server, which can be very costly.
Here, we directly replace the transition function $\mathcal{T}$ with the world model $M_{w}$. Specifically, the next synthetic observation is generated with:

\begin{equation} \label{equation:wm_synthetic_observation}
\hat{o}^t \sim \pi_{\theta_w}(\cdot|I_w, c^w_t)
\end{equation}

Then, in the next step, the policy model generates next action conditioned on the synthetic observation:

\begin{equation} \label{equation:wm_synthetic_action}
    (\hat{h}_t, \hat{a}_t) \sim \pi_{\theta_M}(\cdot \mid I, q, \hat{c}_t^{\text{clip}}).
\end{equation}

Those collected trajectory is thus $\hat{\tau} = \{(o_0, a_0), (\hat{o}_1, \hat{a}_1), \dots, (\hat{o}_t, \hat{a}_t)\} $, which ultimately forms a trajectory dataset $D_w$ after rejection sampling. By combining $D_i$ from self-improvement and $D_w$, we can get an augmented new training dataset to train a new policy model, WebEvolver.

\paragraph{Inference-time Look-ahead}

To enhance the planning ability during inference, 
we propose a look-ahead mechanism that simulates $d$-step trajectories using both the agent policy model $M_i$ and the world model $M_w$. 
We call this method \textbf{W}orld \textbf{M}odel \textbf{L}ook-\textbf{A}head (WMLA).
For each candidate action $a_t$ at step $t$, 
we first simulate trajectories by generating $d$-step rollouts $\hat{\tau}_w$ through iterative application of:
\vspace{-1em}

\begin{align} \label{eq:WMLA}
    \hat{o}_{t+j} \sim \pi_{\theta_w}(\cdot | I_w, c^w_{t+j}), \quad  \nonumber \\
    (\hat{h}_{t+j}, \hat{a}_{t+j}) \sim \pi_{\theta_M}(\cdot | I, q, \hat{c}^\text{clip}_{t+j}),
\end{align}

where $j\in \{1, \dots, d\}$, $c^w_{t+j}$ and $\hat{c}^\text{clip}_{t+j}$ are truncated histories from the world model and policy model, respectively. 

Next, we evaluate trajectories by employing an LLM-based evaluator to score each rollout $\hat{\tau}_w$. Following \citet{AgentTreeSearch, webdreamer}, the evaluator assigns a scalar from $\{0, 0.5, 1.0\}$ (incorrect, on track, or complete) based on the trajectory's alignment with task completion. 
Finally, we select the optimal action $a^*_t = \arg\max_{a_t} \text{Score}(a_t)$ that maximizes expected progress.

\section{Experiments}

\begin{table*}[t]
\vspace{0.1in}
\centering
\small
\renewcommand{\arraystretch}{1.1}
\setlength{\tabcolsep}{1.mm}{
\scalebox{0.98}
{\begin{tabular}{@{}l@{}ccccccccccc@{}|cc@{}}
\toprule
& \multirow{2}{*}{ \tabincell{c}{ AllRe-\\cipes}} & \multirow{2}{*}{ \tabincell{c}{ Apple}} & \multirow{2}{*}{ \tabincell{c}{ ArXiv}} & \multirow{2}{*}{ \tabincell{c}{ BBC}} & \multirow{2}{*}{ \tabincell{c}{ Cam\\Dict}} & \multirow{2}{*}{ \tabincell{c}{ Cour-\\sera}} & \multirow{2}{*}{ \tabincell{c}{ \footnotesize{ESPN}}}   & \multirow{2}{*}{ \tabincell{c}{ Git\\Hub}}   & \multirow{2}{*}{ \tabincell{c}{ Google\\Map}} & \multirow{2}{*}{HF} &  \multirow{2}{*}{ \tabincell{c}{ Wolfram\\Alpha}} & \multirow{2}{*}{\tabincell{c}{\textbf{WV}\\ \textbf{All}} } & \multirow{2}{*}{\tabincell{c}{\textbf{M2W}\\ \textbf{Live}} }   \\ 
\\ 
\hline
GPT-4o-mini & 44.44  & 39.53 & 23.26 & 21.43  & 30.23 & 35.71 & 27.27 & 31.71 & 41.46  & 25.58 & 36.96 & 32.55 & 16.98 \\
GPT-4o & 31.11  & 41.86 & 27.91 & 32.56  & 41.86 & 47.62 & 27.27 & 36.59 & 36.58  & 46.51 & 56.52 & 38.83 & 20.75 \\
\hline 
\multicolumn{2}{@{}l}{\textit{Self-Improving}} \\
Llama-3.3 70B & 35.56 & 39.53 & 9.30 & 28.57  & 37.21 & 38.10 & \underline{50.00}& 24.39& 34.15& 23.26& 41.30 & 32.98 & 18.86 \\
self-improve (1) & 55.56 & 39.53 & 27.91 & 45.24 & 20.93 & \underline{61.90} & 34.09 & 39.02 & 39.02 & 23.26 & 39.13 & 38.68 & 15.09 \\
self-improve (2) & 40.00 & 30.23 & 27.91 & 30.95 & 32.56 & 59.52 & 29.55 &  43.90 & 46.34 & \underline{41.46} & 39.13 & 38.23 & 16.98 \\
self-improve (3) & 44.44 & 30.23 & 32.25 & 33.33 & 32.56 & 47.62 & 31.81 &  \underline{43.90} & \underline{48.78} & 34.89 & 45.65 & 38.65 & 16.98 \\
Synthetic Traj. &  55.56 & \underline{41.86} & 32.25 & 35.71 & 34.89 & 46.51 & 31.81 &  34.14 & 36.59 & 34.89 & 43.47 & 38.98 & 18.86 \\
{\bf \textbf{WebEvolver}} & \underline{62.22} &30.23&\underline{37.21}&\underline{47.62} & \underline{53.49} & {59.52} & 34.09 & 26.83 & 46.34 & 23.26 & \underline{45.65} & \underline{42.49}  & \underline{22.64}  \\
\hline
\multicolumn{2}{@{}l}{\textit{Inference-time Look-ahead}} \\
+ WebDreamer & 64.44  & 41.86 & 44.19 & \textbf{57.14} & 30.23 & 59.52 & 20.45  & 41.46 & 46.34  & 41.86 & 43.48 & 44.61 & 22.64  \\
+ {\bf \textbf{WMLA}} ($d$=1) & \textbf{66.67}  & \textbf{46.51} & 39.53 & {42.86} & 32.56 & \textbf{69.05} & 22.73 & 43.90 & \textbf{68.29}  & 37.21 & 41.46 & 46.24 & \textbf{28.30} \\
+ {\bf \textbf{WMLA}} ($d$=2) & 64.44 & 41.86 & \textbf{46.51} & {42.86} & \textbf{62.79} & 66.67  & \textbf{40.91}  & \textbf{46.34} & 43.90 & \textbf{53.49} & \textbf{54.34} & \textbf{51.37} & 24.53 \\

\bottomrule 
\end{tabular}}}
\caption{Task success rate on Text-only WebVoyager test set (WV; 473 queries) and Mind2Web-Live-filtered test set (M2W Live; 53 queries). {\bf WebEvolver} and {\bf WMLA} are our approaches. For \textit{Inference-time Look-ahead}, the backbone policy model we use is WebEvolver. We leave more inference-time look-ahead results on different policy models in Figure~\ref{fig:self_improve_results}. \underline{Underline} indicates the best among self-improving, and \textbf{bold} indicates the best performance when inference-look ahead is applied. }  
\label{tab:main_result}
\end{table*}

\subsection{Setup}

We use the Cognitive Kernel~\citep{cognitive_kernel} as the foundation agent framework, specifically its Web Agent Module for autonomous Web interaction. 
Here, the state space $\mathcal{S}$ is the whole Internet, powered by Playwright\footnote{A Javascript version \url{https://playwright.dev}} in the Web docker in Cognitive Kernel.
The action space include \texttt{type}, \texttt{click}, \texttt{scroll}, \texttt{goback}, \texttt{stop}, and \texttt{restart}. 
At each time step $t$, the observation $o_t$ is the accessibility tree of the visible components in the virtual browser, simulating what humans can perceive when browsing online. 
The transition function $\mathcal{T}$ executes atomic browser actions based on the current webpage state, updates the webpage, and thus the observation accordingly, 
and handles execution errors by feeding them back to the reasoning system until task completion or step limit is reached.
Regarding the evaluation protocol $\mathcal{R}$, we address potential false negatives in human-annotated step-wise comparisons~\citep{webcanvas} by employing GPT-4o for end-to-end task completion assessment, following the methodology of \citet{webvoyager}. 
This method accommodates the existence of multiple distinct trajectories that can each successfully accomplish the same task objective, other than the human-annotated ones.
GPT-4o will be provided the full trajectory of the task and asked to evaluate whether the original query $q$ is completed or not, yielding a binary score of 0 or 1.

Regarding self-improvement, the backbone agent foundation model $\mathcal{M}$ we use is \texttt{Llama-3.3-70b}, and subsequently the self-improving experiments are also based on \texttt{Llama-3.3-70b}. During rejection sampling, \texttt{Llama-3.3-70b} instead of GPT-4o is used to evaluate whether the task has successfully completed or not.
More details regarding the agent system, including definitions of the atomic operations, system prompts, are detailed in Appendix~\ref{app:agent_details}.


We select two live web navigation benchmarks for experiments, WebVoyager~\citep{webvoyager} and Mind2Web-Live~\citep{webcanvas}. 
Here, the web agent is expected to interact with the real-world web environment to complete the task.
Since some websites are not accessible in our experimental web environment, either due to geographical locations or IP blocks, we filter out some websites for our experiments\footnote{Details about the websites are presented in Appendix~\ref{app:website_filter}}.
To ensure robustness, we conduct our experiments roughly at the same time window twice and report the average results.


\subsection{Self-Improvement}

We use \texttt{Llama3.3-70B} as the backbone LLM $\mathcal{M}$ for sampling and self-improving.
For the training query, we follow OpenWebVoyager~\citep{openwebvoyager}\footnote{\url{https://github.com/MinorJerry/OpenWebVoyager/tree/main/WebVoyager/data_for_training/IL}} to use the training set of Mind2web and self-instructed queries from both the websites in WebVoyager and Mind2web, in total 1,516 queries.
We first use \texttt{Llama3.3-70B} as the backbone agent policy model for sampling queries, and conduct a round of rejection sampling using \texttt{Llama3.3-70B} itself as the backbone for evaluation function $\hat{r}$\footnote{In the original OpenWebVoyager paper, GPT-4o serves as the backbone for the scoring function. In this work, to ensure a purely self-improving process, we only employ \texttt{Llama3-70B} within the self-improvement loop.}, using the evaluation prompt in Appendix~\ref{app:agent_details}.
The trajectories are then used to fine-tune \texttt{Llama3.3-70B} to acquire the model named \textit{self-improve (iter 1)}. 
Then, we use the improved model to conduct another round of trajectory sampling, where the newly sampled finished trajectories are added to the training data in the first round, to train a new model named \textit{self-improve (iter 2)}. 
In the meantime, we convert the trajectories to the form of training a world model, meaning predicting the next observation $o_t$ based on the scheduled observation $a_{t-1}$ and the histories of the observations.

\begin{figure}[t]
    \centering
    \begin{minipage}[t]{0.24\textwidth}
        \centering
        \includegraphics[width=0.95\linewidth]{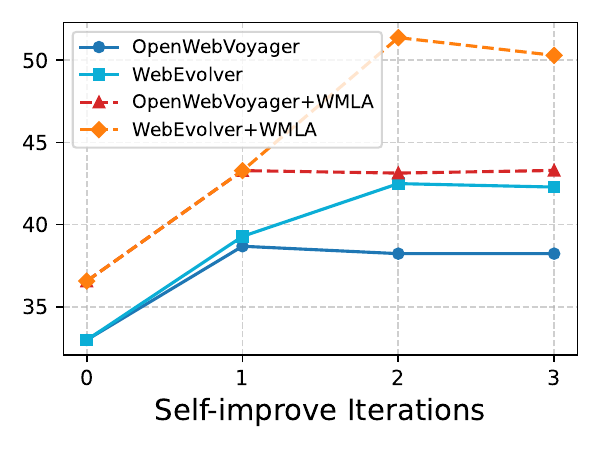}
        \textbf{\small{WebVoyager}}
        \label{fig:webvoyager_si}
    \end{minipage}%
    \hfill
    \begin{minipage}[t]{0.24\textwidth}
        \centering
        \includegraphics[width=0.95\linewidth]{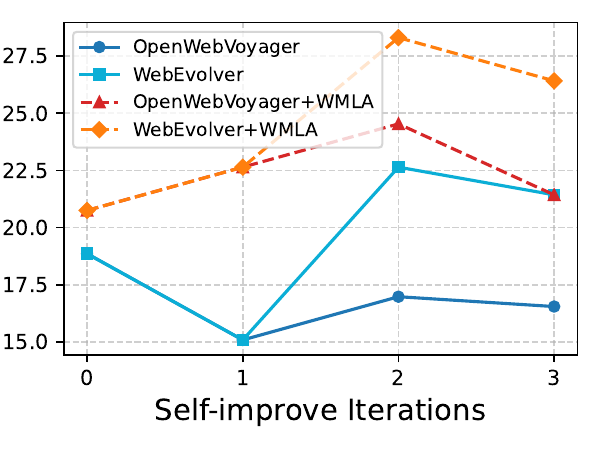}
        \textbf{\small{Mind2web-Live}}
        \label{fig:mind2web_si}
    \end{minipage}
    \caption{Visual illustration of overall success rate evolving on WebVoyager and Mind2Web-Live.}
    \label{fig:self_improve_results}
\vspace{-1em}    
\end{figure}

\paragraph{World Model} 
We adopt a \texttt{Llama3.3-70B} to fine-tune the world model, alongside the self-improving of policy model, to get \textit{world model (iter 1)} and \textit{world model (iter 2)}. 
For synthetic trajectory generation, we use the world model $M_w$ (at iteration 2) and policy model $M_1$ (at iteration 1, which has a better performance). 
For each query  $q$, beginning with an initial observation-action pair $(o_0, a_0)$, we alternate between world model prediction and policy decisions: at each timestep $t$, the world model generates the next synthetic observation $\hat{o}_t$ according to Equation~(\ref{equation:wm_synthetic_observation}), which the policy model then uses to produce the subsequent action $\hat{a}_t$ via Equation~(\ref{equation:wm_synthetic_action}). 
This interaction forms complete synthetic multi-step trajectories $\hat{\tau}$ of length $T=7$ steps, with early termination if the world model generates a terminal state. 
An example if presented in Figure~\ref{fig:example_wm_traj}.
To have a more diverse training set, we only use the queries that are not successfully executed in self-improving iterations to acquire synthetic trajectories.
We apply another round of rejection sampling using the evaluation protocol $\mathcal{R}$, while using zero-shot \texttt{Llama3.3-70B} as the backbone language model to follow the setting of self-improving.
In the end, the world-model-synthesized data are combined with the SFT data in self-improvement, to train \texttt{Llama3.3-70B} to acquire the final model of WebEvolver.

\subsection{Inference-time World Model Look-ahead (WMLA)}

To perform WMLA, we use the policy model $\mathcal{M}$ to sample up to 3 actions. 
At time step $t$, with observation $o_t$,
we use the original policy model with temperature equal to 0 to generate the first action, $a_t^{(1)}$.
Since the fine-tuned policy model will have a sharp output distribution, making it hard to directly sample different actions during decoding, besides setting the decoding temperature to 0.7, we add a sentence of additional prompt to guide the policy model to generate the $k$-th action: \textit{Please generate actions different from} $\{a_t^{(j)}, j \in \{1, \dots, k-1\}\}$.
Then, we use the final world model \textit{world model (iter 2)} and the policy agent model to iteratively sample future look-ahead trajectories based on Equation~(\ref{eq:WMLA}), with a look-ahead depth of 1, 2, and 3.
Then, following WebDreamer, we use GPT-4o as the scoring function to rate each action based on the look-ahead results and choose the action with the highest score for execution.

\begin{table*}[t]
\vspace{0.1in}
\centering
\small
\setlength{\tabcolsep}{1.3mm}{
\scalebox{0.92}{
\begin{tabular}{l|ccc||ccc|ccc|ccc|ccc}
\toprule
\multirow{2}{*}{\textbf{Model}} & \multicolumn{3}{c||}{All} & \multicolumn{3}{c|}{Depth=1} & \multicolumn{3}{c|}{Depth=2} & \multicolumn{3}{c|}{Depth=3} & \multicolumn{3}{c}{Depth$\geq$4} \\
                                & STR & Sim. & O/A & STR & Sim. & O/A & STR & Sim. & O/A & STR & Sim. & O/A & STR & Sim. & O/A \\
\midrule
{\texttt{gpt-4o}} & 40.62 & 33.26 & 37.85 & 41.24 & 35.73 & 40.21 & 38.20 & 32.58 & 36.70 & 36.99 & 31.96 & 37.44 & 42.41 & 32.91 & 37.45   \\
{\texttt{Llama-3.3-70b}} &  39.04 & 32.25 & 38.77 & 43.64 & 39.51 & 34.83 & 39.33 & 34.83 & 41.95 & 39.73 & 33.33& 41.55 & 36.85 & 27.99 & 35.16 \\
{\quad \texttt{iter-1}} & 49.23 & 37.83 & 43.15 & 55.44 & 44.91 & 50.52 & 53.03 & 39.77 & 46.59 & \textbf{53.70} & \textbf{40.28} & \textbf{46.30} & 43.76 & 33.33 & 37.73 \\
{\quad \texttt{iter-2}} & \textbf{56.79} & \textbf{44.77} & \textbf{51.82} & \textbf{75.96} & \textbf{63.56} & \textbf{72.86} & \textbf{57.80} & \textbf{45.14} & \textbf{52.32} & {51.24} & 35.82 & 45.27 & \textbf{50.54} & \textbf{39.94} & \textbf{45.31}  \\
\bottomrule
\end{tabular}
}
\caption{Performance of intrinsic evaluation of world modeling. \textbf{Structural correctness (STR)} measures syntactic validity of the generated accessibility tree,  \textbf{Similarity (Sim.)} assesses alignment with ground-truth webpage content, and  \textbf{Overall assessment (O/A)} evaluates functional and semantic coherence. All values are percentages (range 0-100).
Details of the evaluation metrics ae presented in Section \ref{sec:wm_intrinsic_eval}.
}
\label{tab:world_model_intrinsic}}
\end{table*}


\subsection{Results and Analysis}

In this subsection, we provide results of self-improvements, the effect of WMLA, the intrinsic evaluation of world models, and additional experiments on GAIA.

\paragraph{WebEvolver and WMLA Main Results}

Our key findings are presented in Table~\ref{tab:main_result}, with the progression of self-improvement across iterations visualized in Figure~\ref{fig:self_improve_results}. 
The first two rows of the table establish reference performance using GPT-4o and GPT-4o-mini as foundation models.
In terms of self-improvement,
the initial self-improvement iteration yields a 6\% success rate increase over the zero-shot baseline on WebVoyager, due to enhanced format compliance and task familiarity.
Performance plateaus at iteration 2, suggesting limited gains from additional similar trajectories. 
However, incorporating world-model-synthesized data with iteration 1's supervised fine-tuning (SFT) data produces a further 4\% improvement. 
This has better improvement compared to the baseline approach adapted from \citet{LLM_agent_can_self_improve} that generates synthetic trajectories without world modeling.

For inference-time action selection with WebEvolver, we benchmark against WebDreamer using GPT-4o for both outcome prediction and action evaluation. Our World Model-based Look-ahead (WMLA) demonstrates optimal performance at depth $d=2$, balancing prediction accuracy against computational overhead. Notably, increasing to $d=3$ provides diminishing returns, consistent with our world model's performance characteristics (see Table~\ref{tab:world_model_intrinsic}).

\paragraph{World Model Intrinsic Evaluation}\label{sec:wm_intrinsic_eval}

We evaluate our world model's ability to generate plausible next webpages through three metrics: 
\textbf{Structural correctness (STR)} measuring syntactic validity of the generated accessibility tree, 
\textbf{Similarity (Sim.)} assessing alignment with ground-truth webpage content, and 
\textbf{Overall assessment (O/A)} evaluating functional and semantic coherence.
While real-time information (e.g., from BBC or Hugging Face) inevitably causes hallucinations during generation, 
we do not directly evaluate the degree of hallucination. 
Hallucinations are implicitly captured through Sim.\ and O/A scores, yet they pose minimal risk in our framework. 
In fact, they may enhance diversity and knowledge in synthesized trajectories, with benefits empirically validated by downstream performance gains.
We use GPT-4o to perform an automatic evaluation of all three metrics and normalize the scores to 0$\sim$ 1. The prompt we used is presented in Appendix~\ref{app:agent_details}.
The results are presented in Table~\ref{tab:world_model_intrinsic}.
We can see that the performance degrades sharply (scores $< 0.50$) for generation depths $> 2$, which is in line with the experiments in WMLA that the performance gain diminishes when WMLA depths $\geq 3$.

\begin{table}
\centering
\small
\begin{tabular}{lccc}
\toprule
Model & \tabincell{c}{GAIA \\ Level 1} & \tabincell{c}{GAIA \\ Level 2} & SimpleQA \\
\midrule
Llama 3.3-70b & 19.2 & 10.9 & 36 \\
iter 1 & 26.9 & 15.6 & 44\\
iter 2 & 26.9 & 12.5 & 45\\
WebEvolver & 30.7 & \textbf{17.2} & {48}\\
+ WMLA & \textbf{34.6} & \textbf{17.2} & \textbf{58}\\
\bottomrule
\end{tabular}
\caption{GAIA-web and SimpleQA performance.}
\label{tab:gaia_web}
\end{table}

\paragraph{Out-of-domain Generalization}
We evaluate our improved agent foundation model on GAIA~\citep{GAIA}, focusing on the web-dependent query subset (GAIA-web)\footnote{\url{https://github.com/MinorJerry/WebVoyager/blob/main/data/GAIA_web.jsonl}}, and also SimpleQA~\cite{simpleqa}, where we use web agent to explore the answers.
Since GAIA typically require multi-step web navigation combined with arithmetic/logical reasoning. 
and the self-improved agent LLM focuses solely on action generation, we adopt a hybrid approach: we use GPT-4o to decompose queries into sub-tasks that web agents can address, and also leverage GPT-4o for result generation and calculation. 
The web agent component is based on Llama-based models including WebEvolver.
We use bing.com instead of Google due to CAPTCHA challenges, which can also demonstrating our method's out-of-domain generalization since the training data does not contain trajectories in bing.com. 
Results on Table~\ref{tab:gaia_web} show consistent improvement on Level 1 and SimpleQA queries through self-improvement and world model augmentation, mirroring trends observed in WebVoyager and Mind2web-live. 
However, Level 2 queries, which demand deeper reasoning and extended multi-step interactions, show limited gains, as these capabilities lie beyond our current training scope. 
This limitation highlights an important direction for future work in developing agents for complex, real-world web tasks.

\begin{table}[h]
\centering
\small
\begin{tabular}{lc}
\toprule
$k$ & WebVoyager \\
\midrule
2 & 48.62 \\
3 & \textbf{51.37} \\
5 & 50.73 \\
\bottomrule
\end{tabular}
\caption{Ablations on the branching factor $k$ in WMLA.}
\label{tab:gaia_web}
\end{table}

\paragraph{Ablations on the Branching Factor $k$} We conducted explicit ablation on the branching factor $k$ (number of sampled candidate actions) in WMLA. Performance plateaus around $k$=3-5 because: possible action spaces rarely yield >5 substantially different options per state. $k$=3 as in the paper is still the best choice.

\paragraph{Analysis of World-Model Synthesized Trajectories}
We provide two cases on the world-model synthesized trajectories, indicating that LLM itself contains useful knowledge about the common structures of the web and has the potential to provide diverse trajectories.
It is provided in Figure~\ref{fig:example_wm_traj}.
This case demonstrates an operation involving a click on the `sort by` menu in the GitHub search console. 
Although the world model has not been further fine-tuned on trajectories that include clicking the `sort by` button, it is still able to accurately generate the menu items for GitHub Search, such as sorting by best match, most stars, and so on. 
This capability arises from the commonsense knowledge inherently encoded in the LLM. 
We find that this feature is highly beneficial for improving the diversity of interactions with previously unseen websites.

\section{Conclusion}

In this paper, we present WebEvolver, a framework for agent foundation model self-improvement through co-learning with a world model, which enhances the effectiveness of the self-improvement cycle. The co-learned world model can also be utilized for inference-time look-ahead, aiding in the selection among different sampled actions.
Experiments on WebVoyager, Mind2Web-Live, and GAIA-web demonstrate the effectiveness of boosting the performance of self-improving agent.

\section*{Limitations}

First, the agent system we use includes only an action generation module, whereas recent studies have shown that incorporating a standalone planning module can further enhance agent performance. However, planning is orthogonal to our research focus.
Second, because we focus on open-domain, real-world web environments, websites may change over time, making it difficult for future work to exactly replicate the same web conditions. To ensure fair comparisons in our experiments, we complete all tasks within approximately the same time frame. Additionally, we include GAIA-web and SimpleQA as two supplementary evaluation datasets, as they primarily focus on factual questions and are less susceptible to significant changes over time.

\bibliography{agent_ref, agent_self_improve, ref, custom}

\appendix


\section{Details of Agent Implementation}\label{app:agent_details}




In this section, we present additional details of the prompt we used for the web agent.

The system prompt for web agent action generation:

\onecolumn
\begin{tcolorbox}[colback=blue!5!white, colframe=blue!75!black, breakable, title = {\textsc{Agent System Prompt}}]\label{prompt:wm_inherent}

\ttfamily
\small

You are an autonomous intelligent agent tasked with navigating a web browser. You will be given web-based tasks. These tasks will be accomplished through the use of specific actions you can issue.\\

Here's the information you'll have:\\
\begin{itemize}[leftmargin=*]
    \item The user's objective: This is the task you're trying to complete.
    \item The current observation (web page's accessibility tree): This is a simplified representation of the webpage, providing key information. Optionally, you may be provided with a screenshot of the webpage. You should pay close attention to the screesnhot to make decisions.
    \item The open tabs: These are the tabs you have open.
    \item The previous actions: You can refer to the conversation history with the user to see the actions you have taken. It may be helpful to track your progress.
\end{itemize}

The actions you can perform are the following:
\begin{itemize}[leftmargin=*]
  \item `click [id]`: This action clicks on an element with a specific id on the webpage.
  \item `type [id] [content] [press\_enter\_after=0|1]`: Use this to type the content into the field with id. By default, the \"Enter\" key is pressed after typing unless press\_enter\_after is set to 0.
  \item `wait`: Wait for the page to load, with a duration of 5 seconds.
  \item `goback`: Navigate to the previously viewed page.
  \item `restart`: Navigate to the Google search homepage. When you can't find information in some websites, try starting over from Google search.
  \item `stop [answer]`: Issue this action when you believe the task is complete. If the objective is to find a text-based answer, provide the answer in the bracket. If you believe the task is impossible to complete, provide the answer as "N/A" in the bracket.
\end{itemize}

To be successful, it is very important to follow the following rules:\\

1. You should only issue an action that is valid given the current observation. For example, you should NOT type into buttons or click on statictext.\\

2. You should only issue one action at a time.\\

3. STRICTLY Avoid repeating the same action if the webpage remains unchanged. You may have selected the wrong web element or numerical label. Continuous use of the Wait is also NOT allowed.\\

4. Issue stop action when you think you have achieved the objective. Don't generate anything after stop.\\

Your reply should strictly follow the format:
Thought: \{\{Your brief thoughts (briefly summarize the info that will help complete the task)\}\} Action: ```\{\{the next action you choose to take\}\}```

\end{tcolorbox}

The system prompt for using world model as a web server, by generating the next observation based on current observation and the scheduled action. We present two variation of world model objectives, the first one is to only predict an abstract short description of what the next observation is (denoted as \textbf{Abstract Description}), and the second one is to predict the structured accessibility tree of the next observation (denoted as \textbf{Accessibility Tree)}.

\begin{tcolorbox}[colback=blue!5!white, colframe=blue!75!black, breakable, title = {\textsc{World Model Look-Ahead (Abstract Description)}}]\label{prompt:wm_inherent}

\ttfamily
\small

You are a web server. You are given the current observed accessibility tree of the web page, and an action to perform. \\

The expected output is a short description on what the next observation is, in the form of free text.

The definitions of the actions are as follows: The actions you can perform are the following:
\begin{itemize}[leftmargin=*]
  \item `click [id]`: This action clicks on an element with a specific id on the webpage.
  \item `type [id] [content] [press\_enter\_after=0|1]`: Use this to type the content into the field with id. By default, the \"Enter\" key is pressed after typing unless press\_enter\_after is set to 0.
  \item `scroll [direction=down|up]`: Scroll the page up or down.
  \item `goback`: Navigate to the previously viewed page.
  \item `restart`: Navigate to the original home page and restart the action.
\end{itemize}
\end{tcolorbox}


\begin{tcolorbox}[colback=blue!5!white, colframe=blue!75!black, breakable, title = {\textsc{World Model Look-ahead (Accessibility Tree)}}]\label{prompt:wm_inherent}

\ttfamily
\small
You are an intelligent assistant designed to interact with web pages through an accessibility tree. Your task is to predict the accessibility tree of the next web page based on the given starting accessibility tree and a specified action.
The format of accessibility tree:\\

Tab 0 (current): Google \textbackslash n \textbackslash n[1] RootWebArea 'Google' focused: true\textbackslash n\t[2] link 'Gmail '\textbackslash n\t[3] link 'Search Image '\textbackslash n\t[4] button 'Google Apps' expanded: false\textbackslash n\t[5] link 'Log in'\textbackslash n\t[6] image '2024'\textbackslash n\t[7] combobox 'Search' focused: true autocomplete: both hasPopup: listbox required: false expanded: false\textbackslash n\t[8] button 'Share'\\

The format of action:\\

type [7] [JQuery selector for elements with specific class] [1]\\

which indicates typing "JQuery selector for elements with specific class" into the field with id 7, corresponding to the combobox (search box) on the Google homepage.\\

The definitions of the actions are as follows: The actions you can perform are the following:\\

\begin{itemize}[leftmargin=*]
  \item `click [id]`: This action clicks on an element with a specific id on the webpage.
  \item `type [id] [content] [press\_enter\_after=0|1]`: Use this to type the content into the field with id. By default, the \"Enter\" key is pressed after typing unless press\_enter\_after is set to 0.
  \item `scroll [direction=down|up]`: Scroll the page up or down.
  \item `goback`: Navigate to the previously viewed page.
  \item `restart`: Navigate to the Google search homepage. When you can't find information in some websites, try starting over from Google search.
\end{itemize}
\end{tcolorbox}

The system prompt for automatic evaluation of a web agent task.

\begin{tcolorbox}[colback=blue!5!white, colframe=blue!75!black, breakable, title = {\textsc{Automatic Evaluation}}]\label{prompt:wm_inherent}

\ttfamily
\small

As an evaluator, you will be presented with three primary components to assist you in your role:\\

1. Web Task Instruction: This is a clear and specific directive provided in natural language, detailing the online activity to be carried out. These requirements may include conducting searches, verifying information, comparing prices, checking availability, or any other action relevant to the specified web service (such as Amazon, Apple, ArXiv, BBC News, Booking etc).\\

2. Result Webpage Accessibility Tree: This is a representation of the web page showing the result or intermediate state of performing a web task. It serves as proof of the actions taken in response to the instruction.\\

3. Result Response: This is a textual response obtained after the execution of the web task. It serves as textual result in response to the instruction.\\

\begin{itemize}[leftmargin=*]
  \item You DO NOT NEED to interact with web pages or perform actions such as booking flights or conducting searches on websites.
  \item You SHOULD NOT make assumptions based on information not presented in the webpage when comparing it to the instructions.
  \item Your primary responsibility is to conduct a thorough assessment of the web task instruction against the outcome depicted in the screenshot and in the response, evaluating whether the actions taken align with the given instructions.
  \item NOTE that the instruction may involve more than one task, for example, locating the garage and summarizing the review. Failing to complete either task, such as not providing a summary, should be considered unsuccessful.
  \item NOTE that the screenshot is authentic, but the response provided by LLM is generated at the end of web browsing, and there may be discrepancies between the text and the screenshots.
  \item Note the difference: 1) Result response may contradict the screenshot, then the content of the screenshot prevails, 2) The content in the Result response is not mentioned on the screenshot, choose to believe the content.
\end{itemize}

You should elaborate on how you arrived at your final evaluation and then provide a definitive verdict on whether the task has been successfully accomplished, either as 'SUCCESS' or 'NOT SUCCESS'.

\end{tcolorbox}

The system prompt for automatic evaluation of world modeling.

\begin{tcolorbox}[colback=blue!5!white, colframe=blue!75!black, breakable, title = {\textsc{World Model Intrinsic Evaluation}}]\label{prompt:wm_inherent}

\ttfamily
\small

You are tasked with evaluating the accuracy of ntnerated accessibility tree against a ground truth accessibility tree obtained from an actual web server. Your evaluation should focus on three main criteria: structure correctness, element correctness, and similarity. Follow the instructions below to perform a detailed comparison:\\

Criteria for Evaluation:

1. **Structure Correctness**:
\begin{itemize}[leftmargin=*]
   \item  Ensure that the basic hierarchy and relationships between elements in the generated tree match the ground truth.
   \item Ensure that interactive elements (like buttons, links, forms) are correctly represented and maintain their intended functionality.
\end{itemize}

2. **Similarity (GPT-score)**:
\begin{itemize}[leftmargin=*]
   \item Assess how similar the generated content is compared to the ground truth.
   \item Provide a similarity score based on the overall content and structure comparison.
\end{itemize}

3. **Overall Functionality Assessment**:

\begin{itemize}[leftmargin=*]
   \item Compare the functional coherence of the generated tree to the ground truth tree, focusing on the representation and functionality of interactive elements.
   \item Evaluate the semantic coherence of the generated tree, ensuring that it conveys the same meaning and purpose as the ground truth.
\end{itemize}

For example, if if the webpage is on Allrecipe, as long as the generated tree contain necessary recipe, no matter hallucination, it can be considered as success.
For example, if the webpage is on google, in searching for some information, then only consider whether the generated tree contain roughly necessary information without the need to check the factuality.\\

1. **Input Trees**:
\begin{itemize}[leftmargin=*]
   \item You will be provided with two accessibility trees: one generated by a language model simulating a web browser, and one obtained from an actual web server.
\end{itemize}

2. **Output Format**:\\
- Provide rationale of your findings, including:
\begin{itemize}[leftmargin=*]
     \item Structural discrepancies
     \item Similarity score with an explanation
     \item Scores should be selected from [0, 1, 2, 3]. 3 means exactly the same and 0 means a total failure of generation.
\end{itemize}

\#\#\# Example Output

Structure Correctness: [THOUGHT]\textbackslash n Score: [score]\textbackslash n \\
Similarity: [THOUGHT]\textbackslash n Score: [score]\textbackslash n \\
Overall Functionality Assessment: [THOUGHT]\textbackslash nScore: [score]\textbackslash n 
\end{tcolorbox}

\section{Additional Details on Mind2web-live and WebVoyager Dataset} \label{app:website_filter}

We conduct our evaluations using a subset of the testing portion of Mind2Web-Live\footnote{\url{https://huggingface.co/datasets/iMeanAI/Mind2Web-Live/blob/main/mind2web-live_test_20241024.json}} and WebVoyager\footnote{\url{https://github.com/MinorJerry/WebVoyager/blob/main/data/WebVoyager_data.jsonl}}. 
Here is a list of the websites that are excluded:

\begin{tcolorbox}[colback=red!5!white, colframe=red!75!black, breakable, title = {\textsc{Excluded Websites}}]

\ttfamily
\small

EXCLUDED\_WEBSITES\_MIND2WEB = \{
'exploretock', 'kohls', 'united', 'parking', 'viator', 'delta', 'redbox', 'soundcloud', 'gamestop', 'travelzoo', 'amctheatres', 'ryanair', 'cargurus', 'resy', 'rentalcars', 'kbb', 'cabelas', 'menards', 'yellowpages', 'tripadvisor', 'tiktok.music', 'stubhub', 'thumbtack', 'weather', 'uhaul', 'health.usnews', 'healthgrades', 'theweathernetwork', 'zocdoc', 'usnews.education', 'epicurious', 'osu.edu', 'ups', 'dmv.virginia.gov', 'extraspace', 'finance.yahoo', 'pinterest',
'sixflags', 'spothero', 'justice.gov', 'foxsports', 'ign', 'koa', 'tvguide', 'webmd', 'sports.yahoo', 'babycenter', 'tesla',
\}

EXCLUDED\_WEBSITES\_WEBVOYAGER = \{
'booking', 'espn', 'amazon', 'google', 'googleflight'
\}
\end{tcolorbox}

\begin{figure*}[t]
    \centering
    \includegraphics[width=\textwidth]{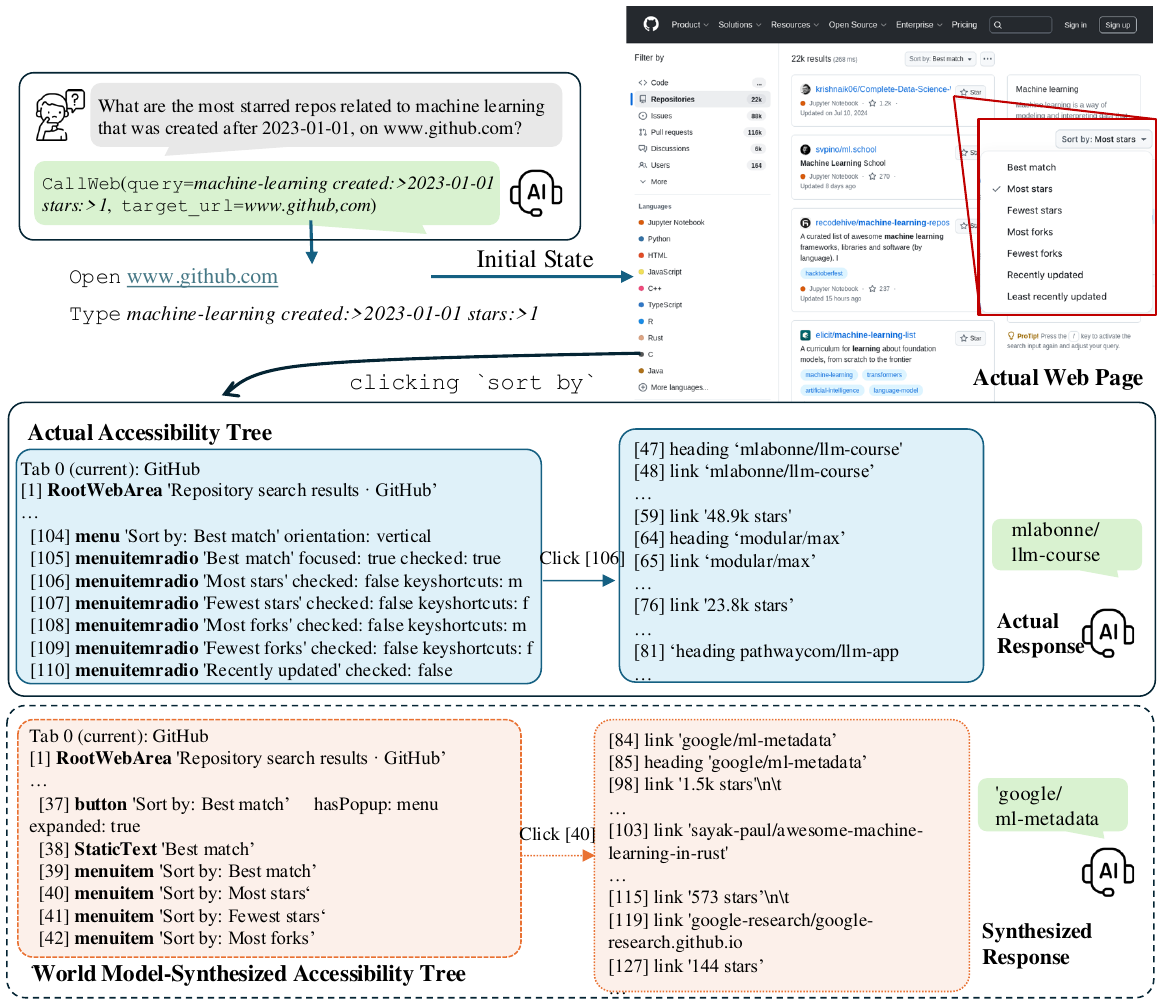} 
    \caption{An example of world model-synthesized trajectory.}
    \label{fig:example_wm_traj}
\end{figure*}

\end{document}